\documentclass[pdflatex,sn-mathphys-num]{sn-jnl}

\usepackage{graphicx}%
\usepackage{multirow}%
\usepackage{amsmath,amssymb,amsfonts}%
\usepackage{amsthm}%
\usepackage{mathrsfs}%
\usepackage[title]{appendix}%
\usepackage{xcolor}%
\usepackage{textcomp}%
\usepackage{manyfoot}%
\usepackage{booktabs}%
\usepackage{algorithm}%
\usepackage{algorithmicx}%
\usepackage{algpseudocode}%
\usepackage{listings}%
\usepackage{longtable}

\usepackage{comment}

\theoremstyle{thmstyleone}%
\theoremstyle{thmstyletwo}%

\theoremstyle{thmstylethree}%

\begin{document}

\title[Article Title]{Predicting High-Risk Colorectal Polyps in African Americans Using Pre-Colonoscopy Clinical Features: Machine Learning Model Development and Temporal Validation}


\author*[1]{\fnm{Basheer} \sur{Qolomany}}\email{Basheer.Qolomany@Howard.edu}

\author[1]{\fnm{Mrinalini} \sur{Deverapall}}\email{Mrinalini.Deverapal@howard.edu}

\author[1]{\fnm{Adeyinka} \sur{Laiyemo}}\email{adeyinka.laiyemo@Howard.edu}

\author[1]{\fnm{Zaki} \sur{Sherif}}\email{zaki.sherif@Howard.edu}

\author[2]{\fnm{Mori} \sur{Yuichi}}\email{yuichi.mori@medisin.uio.no}

\author[3]{\fnm{Omer} \sur{Ahmed}}\email{o.ahmad@ucl.ac.uk}

\author[1]{\fnm{Hassan} \sur{Brim}}\email{hbrim@Howard.edu}

\author[1]{\fnm{Hassan} \sur{Ashktorab}}\email{hashktorab@Howard.edu}

\affil[1]{\orgdiv{Departments of Internal Medicine, Pathology, and Biochemistry, and Cancer Center}, \orgname{Howard University College of Medicine}, \orgaddress{\city{Washington, D.C.}, \postcode{20059}, \country{USA}}}

\affil[2]{\orgdiv{Clinical Effectiveness Research Group}, \orgname{University of Oslo}, \orgaddress{\city{Oslo}, \country{Norway}}}

\affil[3]{\orgdiv{Wellcome/EPSRC Centre for Interventional and Surgical Sciences}, \orgname{University College London Hospital}, \orgaddress{\city{London}, \country{UK}}}


\abstract{
\textbf{Purpose:}

Risk stratification for advanced colorectal polyps typically relies on colonoscopy and/or pathology findings. However, there is growing interest in whether non-invasive features available prior to colonoscopy can help identify patients at higher risk. Such approaches may enhance clinical decision-making by prioritizing surveillance for individuals most likely to harbor high-risk polyps, when colonoscopy resources are limited while potentially reducing unnecessary procedures in lower-risk patients. Importantly, the use of non-invasive, pre-procedural information may also help promote more equitable access to risk stratification, particularly in settings where colonoscopy resources are limited or unevenly distributed. We aimed to develop and externally validate machine learning models to predict high-risk colorectal polyps using only non-invasive, pre-colonoscopy demographic, clinical, and behavioral features in a diverse, predominantly African American, urban cohort.

\textbf{Methods:} 

We conducted a retrospective cohort study using demographic, lifestyle, and comorbidity data from patients who underwent colonoscopy at Howard University Hospital to develop and validate several machine learning models, including neural networks, random forest, support vector machines (SVM), Naïve Bayes, logistic regression, decision trees, k-nearest neighbors (KNN), and XGBoost, for predicting high-risk colorectal polyps. High-risk polyps (HRP) were defined as villous or tubullovillous adenomas, high-grade dysplasia, polyps $\geq$ 10 mm in size, and/or the presence of $\geq$ 3 polyps per procedure; all other cases were classified as low-risk polyps (LRP). 
The dataset included 4,681 patients from 2015-2022 used for internal validation and 1,562 patients from 2023-2024 used for external validation. Model performance was evaluated using the area under the receiver operating characteristic curve (ROC-AUC), precision-recall area under the curve (PR-AUC), accuracy, precision, recall, and F1 score. Model interpretability and feature contribution were assessed using SHapley Additive exPlanations (SHAP).

\textbf{Results:} 

Overall predictive performance was moderate using non-invasive pre-colonoscopy features. The neural network demonstrated the strongest overall discrimination, achieving the highest internal validation performance (ROC-AUC 0.78, PR-AUC 0.75, accuracy 0.72), but showed reduced performance in the external cohort (ROC-AUC 0.67, accuracy 0.66), suggesting potential overfitting or temporal feature drift. In contrast, simpler models including Naïve Bayes, SVM, and XGBoost exhibited lower internal performance (ROC-AUC 0.54-0.59) but more stable generalization to the external cohort (ROC-AUC 0.52-0.63; accuracy approximately 0.53-0.60).
Model interpretability analysis using SHAP identified age, smoking status, sex, occupation, race, colonoscopy indication, and family history of colorectal cancer as the most influential predictors, highlighting contributions from both traditional clinical and sociodemographic factors.

\textbf{Conclusions:}

Prediction of HRP using routine pre-colonoscopy data is feasible but demonstrates limited generalizability across cohorts. These findings highlight the clinical potential and limitations of pre-procedural risk modeling, especially in diverse, underserved populations. Integration of additional data modalities may be required to achieve clinically robust, and equitable prediction tools.

}




\maketitle

\section{Introduction}\label{sec1}

Colorectal cancer (CRC) remains one of the leading causes of cancer-related morbidity and mortality worldwide, with disproportionate impacts on certain racial and socioeconomic groups, despite significant advances in screening and prevention strategies \cite{siegel_colorectal_2023, gupta_recommendations_2020}. Most CRCs arise from adenomatous polyps through the adenoma-carcinoma sequence, which can span over a decade \cite{brenner_protection_2011}. Early identification and removal of these precursor lesions through colonoscopy have substantially reduced CRC incidence and mortality \cite{click_association_2018, rex_colorectal_2017, zauber_colonoscopic_2012}. However, despite the proven effectiveness of colonoscopy, its diagnostic yield is influenced by patient, procedural, and endoscopist-related factors, including bowel preparation quality, withdrawal time, lesion characteristics, and operator variability \cite{adler_interval_2015}.

Traditional risk stratification for colorectal neoplasia relies on broad demographic and clinical variables such as age, sex, family history, and lifestyle factors \cite{sninsky_risk_2022, jeon_determining_2018}. While these factors offer valuable population-level insights, their predictive capacity for individual patient risk remains limited and may miss high-risk patients \cite{koker_machine_2025}. Furthermore, current guidelines for surveillance intervals after colonoscopy are often applied uniformly, without accounting for complex interactions among clinical, behavioral, and procedural variables \cite{urban_deep_2018}, leading to both overuse and underuse of colonoscopy.  Recent advancements in artificial intelligence (AI) and machine learning (ML) have enabled more precise and personalized risk modeling in gastroenterology \cite{chen_accurate_2018, kudo_artificial_2019, kwak_integrative_2025}. Pre-procedure risk prediction has the potential to optimize clinical workflows and improve patient outcomes, particularly in settings where access to invasive testing may be limited. While many AI applications in endoscopy focus on real-time image analysis, ML models that leverage structured demographic and clinical data offer scalable, non-invasive predictive tools that can be applied prior to colonoscopy. However, existing models often lack multidimensional data integration and robust validation in diverse, real-world populations \cite{ba_development_2024}. 

Several colorectal cancer risk prediction models and risk-scoring systems have previously been developed \cite{yeoh_asia-pacific_2011, jeon_determining_2018, sninsky_risk_2022} to support individualized screening and surveillance strategies. Traditional approaches, including the Asia-Pacific Colorectal Screening Score and other questionnaire-based models, commonly incorporate demographic and lifestyle factors such as age, sex, smoking history, BMI, and family history of colorectal cancer to estimate neoplasia risk. While these models have demonstrated moderate predictive performance, they are often based on conventional statistical frameworks and may not fully capture nonlinear interactions among clinical and behavioral variables. In addition, many existing models have been developed in predominantly non-African American populations, limiting understanding of their performance and generalizability in historically underserved groups.

Machine learning approaches may offer advantages \cite{ba_development_2024, topol_high-performance_2019} for pre-colonoscopy risk stratification by integrating multidimensional electronic health record data and identifying complex relationships among routinely collected variables without requiring invasive testing, imaging, or biomarker analysis. Importantly, such models could potentially automate or simplify risk estimation using information already available within clinical records, thereby complementing traditional questionnaire-based risk assessment approaches. However, the incremental benefit of machine learning over simpler risk scoring strategies remains uncertain, particularly in diverse real-world populations.

Despite these advances, few studies have explored the integration of multi-dimensional clinical data; including comorbidities, medication use, social determinants, and procedural features; Such features may capture hidden biological or behavioral contributors to CRC progression, particularly in underrepresented populations, highlighting the need to develop comprehensive models that generalize across diverse groups. Moreover, limited external validation of these models raises concerns about reproducibility and clinical translation, particularly in diverse patient populations.

To address these gaps, we retrospectively analyzed a cohort of 6,243 colonoscopy patients at Howard University Hospital (HUH) between 2015 and 2024, aiming to (1) compare baseline clinical characteristics between internal (pre-2023) and external (2023-2024), temporally distinct cohorts, (2) develop and validate ML models using a temporally external cohort from the same institution for detecting high-risk colorectal polyps using only non-invasive pre-colonoscopy features, and (3) assess performance and main predictors in a predominantly African American, urban cohort-an underrepresented group in CRC AI studies. Our approach leverages diverse EHR data, excludes intra-procedural variables, and is designed to maximize clinical applicability and reduce disparities. In summary, this study, based on a retrospective design and temporal external validation approach, aims to contribute to the growing field of AI-assisted gastroenterology by:
\begin{itemize}
\item Developing and validating robust ML-based models for predicting colorectal polyp and high-risk adenoma risk using pre-procedure clinical data.
\item Conducting internal and external validation across temporally distinct cohorts to assess model generalizability.
\item Providing insights into key demographic and behavioral determinants of polyp risk in a predominantly African American, urban population; an underrepresented group in prior CRC AI studies.
\end{itemize}
\section{Related Work}\label{sec2}
Most prior approaches for colorectal neoplasia risk stratification have relied on conventional clinical risk scores and questionnaire-based prediction models incorporating demographic, behavioral, and family history variables \cite{yeoh_asia-pacific_2011, jeon_determining_2018, sninsky_risk_2022}. Established models such as the Asia-Pacific Colorectal Screening Score and related adenoma risk indices have demonstrated moderate discrimination for identifying individuals at elevated risk for advanced neoplasia. Common predictors across these models include age, sex, smoking status, obesity, and family history of colorectal cancer.

In contrast, fewer studies have explored pre-colonoscopy predictive models that estimate the likelihood of high-risk adenomas or advanced neoplasia using clinical, demographic, and lifestyle data. Traditional statistical risk models, such as the Adenoma Risk Index and Asia-Pacific Colorectal Screening Score, incorporate factors like age, sex, family history, smoking, and BMI to prioritize screening \cite{zauber_colonoscopic_2012, adler_interval_2015, sninsky_risk_2022}. More recently, ML-based approaches using larger and more complex datasets have aimed to improve these tools. Gradient boosting and random forest models have shown moderate performance for predicting adenomas or advanced neoplasia, typically achieving ROC-AUC values between 0.65 and 0.80 when trained on health examination or screening cohorts \cite{koker_machine_2025, urban_deep_2018, chen_accurate_2018}. Despite this progress, external validation frequently demonstrates reduced performance, reflecting degradation, challenges related to population heterogeneity, data drift, and limited external validity \cite{kudo_artificial_2019}.

Recent research has also examined integrating multimodal data sources; including genomics, microbiome composition, and laboratory biomarkers; to improve CRC and polyp risk prediction \cite{kwak_integrative_2025, ba_development_2024, kudo_artificial_2020, gong_edge_2025}. Although promising, these approaches are often resource-intensive and not feasible for routine pre-screening in many clinical settings. Consequently, developing models based solely on routinely available pre-colonoscopy features (e.g., demographic, behavioral, and clinical variables) remains a practical and important goal for precision prevention.

In contrast to prior work that has largely focused on image-based detection or post-procedural predictors, this study advances the literature by developing and temporally validating multiple machine learning models using only non-invasive, pre-colonoscopy features derived from structured electronic health records in a diverse patient population. By integrating demographic, lifestyle, comorbidity, medication, and family history data, our approach enables risk stratification before the procedure is performed, enhancing real-world clinical applicability. Furthermore, model explainability using SHAP analysis provides clinically interpretable insights into key predictors of high-risk colorectal polyps, including age, smoking status, sex, race, occupation, and colonoscopy indication. This emphasis on pre-procedural prediction and temporal generalizability distinguishes our work from prior studies that rely on procedural, pathological, or image-based variables.

\section{Methods}\label{sec3}
\begin{figure}[h]
\centering
\includegraphics[width=1\textwidth]{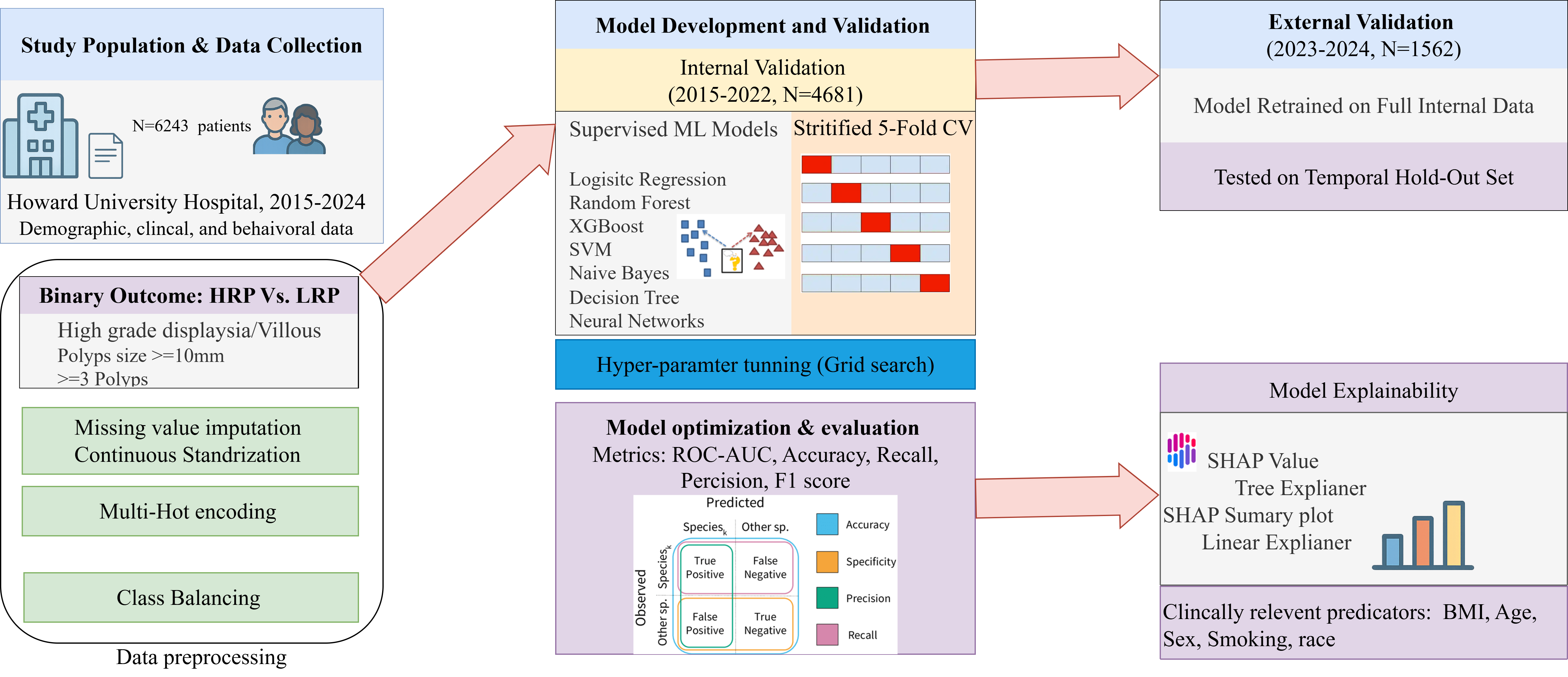}
\caption{Methodology pipeline for HRP predication }\label{fig1_1}
\end{figure}

This study followed a systematic pipeline for data collection, preprocessing, model development, and explainability analysis to predict high-risk polyps (HRP) using demographic, clinical, and behavioral features. The overall workflow is illustrated in Figure \ref{fig1_1}, which outlines each stage of the methodology; from data extraction and feature engineering to model training, validation, and interpretation using SHAP explainability analysis.

\subsection{Study Population}
We conducted a retrospective observational study at Howard University Hospital, encompassing all patients who underwent colonoscopy between January 2015 and Dec. 2024. The dataset included demographic, clinical, and behavioral factors as well as colonoscopy- and pathology-related data. 

After preprocessing, the study cohort comprised 6,243 patients. Data were split a priori based on calendar time rather than randomly to assess temporal generalizability. Patients who underwent colonoscopy between 2015 and 2022 (n = 4,681; 2,018 high-risk polyps [HRP], 2,658 low-risk polyps [LRP]) were used for model development and internal validation, while patients from 2023-2024 (n = 1,562; 769 HRP, 793 LRP) were reserved as an independent temporally held-out validation cohort. This time-based partitioning reflects a real-world deployment scenario and minimizes information leakage across cohorts. The study was conducted under institutional ethical approval, with informed consent waived due to its retrospective design. This study was designed and reported in accordance with the TRIPOD-AI guidelines (see Supplementary Table \ref{secA1}). 

\subsection{Data Collection and Processing}
Clinical and demographic data were extracted from the hospital’s electronic medical record system. Features included demographics (sex, race, age, body mass index [BMI]), lifestyle factors (smoking status, alcohol use, illicit drug use), family history of colorectal cancer in first-degree relatives, comorbidities including coronary artery disease (CAD), diabetes mellitus, hypertension, inflammatory bowel disease (IBD), chronic kidney disease (CKD), dialysis dependence, prior cholecystectomy, Helicobacter pylori infection, and human papillomavirus (HPV) infection, medication exposures (aspirin [ASA], nonsteroidal anti-inflammatory drugs [NSAIDs], multivitamin use, prior chemotherapy or radiation therapy), and colonoscopy indication. All variables used in this study, including behavioral and lifestyle factors such as smoking, alcohol use, and illicit drug use, were extracted from routinely documented electronic health records under Institutional Review Board approval and analyzed in de-identified form in accordance with institutional privacy and data protection policies.

Each patient record was annotated with a binary outcome indicating the presence of high-risk polyps (HRP) or low-risk polyps (LRP), defined according to polyp pathology, size, and number, in alignment with U.S. Multi-Society Task Force (USMSTF) recommendations. A polyp was classified as high-risk if any of the following criteria were met: high-grade dysplasia, villous or tubulovillous histology, largest polyp size $\geq$ 10 mm, or three or more polyps identified during any colonoscopy episode. This definition was applied across the index colonoscopy and up to two prior colonoscopy records when available.

To ensure genuine pre-procedural prediction, only variables available at the time of referral or colonoscopy scheduling were used. Intra- and post-procedural variables (e.g., bowel preparation quality, withdrawal time, polyp location at removal) were intentionally excluded, as these are not accessible for risk stratification before the procedure. This design emphasizes real-world clinical applicability for patient risk triage and early decision support. 

Missing numeric values were imputed using the median strategy, and continuous variables were standardized to zero mean and unit variance using a SimpleImputer and StandardScaler, respectively. To prevent data leakage, imputation and scaling parameters were learned only from the training data within each cross-validation fold and then applied to the corresponding validation data. For external validation, preprocessing was fit on the full internal training cohort and applied unchanged to the external cohort.

For multi-value categorical variables (e.g., Diagnostic Indication), we applied a multi-hot encoding scheme rather than reducing entries to the first code, ensuring no loss of diagnostic information.

Model interpretability was assessed using SHAP, which highlighted clinically relevant predictors, including BMI, age, smoking, alcohol use, sex, and race.
Since high-risk polyps were less frequent, Synthetic Minority Oversampling Technique (SMOTE) was applied within each training fold to balance classes.

\subsection{Model Development and Evaluation}
We implemented a range of supervised learning models for HRP prediction: Logistic Regression, Random Forest, XGBoost, SVM, K-Nearest Neighbors, Naïve Bayes, Decision Tree, and a Neural Network. The neural network was configured with a 6-layer architecture (64-32-16-8-4-2-1), optimized through grid search hyperparameter tuning, and trained using 20 epochs with a batch size of 32, the NN model trained with ReLU activations, sigmoid output, Adam optimizer and binary cross-entropy loss. 

Model training and internal evaluation used stratified 5-fold cross-validation on the 2015-2022 cohort. In each fold, data were imputed, resampled with SMOTE, and used to train the model. Metrics were averaged across folds. For external validation, models were retrained on the entire internal training set and tested on the temporal hold-out set 2023-2024 cohort to assess generalization. 

\subsection{Model Optimization and Evaluation}
Performance was assessed with multiple metrics: ROC-AUC, PR-AUC, accuracy, precision, recall, and F1 score. ROC-AUC and PR-AUC were emphasized because of the class imbalance between HRP and LRP groups. For both internal and external validation, confusion matrices and classification reports were used to quantify sensitivity and specificity.

\subsection{Model Explainability}
For interpretability, we applied SHAP (SHapley Additive exPlanations) to quantify feature’s contribution to HRP prediction across models. Tree-based models (Random Forest, XGBoost, Decision Tree) used TreeExplainer, linear models used LinearExplainer, and the neural network used DeepExplainer. SHAP summary plots highlighted the most influential predictors driving classification decisions, supportingclinical interpretability and confirming the biological plausibility of model-drived risk factors.  
\section{Results}\label{sec4}
\subsection{Basic Characteristics of the Study Population}
From 2015 to 2024, a total of 6,243 patients who underwent colonoscopy at Howard University Hospital were included in the study. Baseline demographic, clinical, and behavioral characteristics are summarized in Table \ref{tab:baseline_characteristics}. The cohort was temporally divided into an internal dataset (2015–2022; n = 4,681, 75\%) and an external dataset (2023-2024; n = 1,562, 25\%). Continuous variables were compared using the Mann-Whitney U test and are reported as mean ± standard deviation.

\begin{table}[htbp]
\caption{Baseline Demographic, Clinical, and Behavioral Characteristics of the Study Cohort. The internal cohort includes patients who underwent colonoscopy between 2015--2022, while the external cohort includes patients who underwent colonoscopy between 2023--2024. Continuous variables are presented as mean $\pm$ standard deviation, and categorical variables are reported as number (percentage).}
\centering
\small
\renewcommand{\arraystretch}{1.2}
\begin{tabular}{|p{3cm}|p{4cm}|p{4cm}|p{1.9cm}|p{1.2cm}|}
\hline
\textbf{Feature} & 
\textbf{Internal (n = 4,681)} & 
\textbf{External (n = 1,562)} & 
\textbf{Test} & 
\textbf{p-value} \\
\hline

Age (years) &
60.16 $\pm$ 9.79 &
60.86 $\pm$ 10.12 &
Mann--Whitney U &
0.0018 \\
\hline

Sex &
0: 2231 (47.7\%); 1: 2450 (52.3\%) &
0: 793 (50.8\%); 1: 769 (49.2\%) &
Chi-square &
0.0358 \\
\hline

Race &
0.0: 4 (0.1\%); 1.0: 3957 (84.5\%); 2.0: 112 (2.4\%); 3.0: 62 (1.3\%); 4.0: 155 (3.3\%); 5.0: 382 (8.2\%) &
0.0: 1 (0.1\%); 1.0: 1310 (83.9\%); 2.0: 54 (3.5\%); 3.0: 18 (1.2\%); 4.0: 68 (4.4\%); 5.0: 111 (7.1\%) &
Chi-square (assumptions violated) &
Not tested \\
\hline

BMI (kg/m$^2$) &
0.0: 111 (2.4\%); 1.0: 948 (20.3\%); 2.0: 1376 (29.4\%); 3.0: 905 (19.3\%); 4.0: 394 (8.4\%); 5.0: 424 (9.1\%) &
0.0: 29 (1.9\%); 1.0: 334 (21.4\%); 2.0: 468 (30.0\%); 3.0: 310 (19.8\%); 4.0: 171 (10.9\%); 5.0: 150 (9.6\%) &
Chi-square &
0.1620 \\
\hline

Smoking &
0.0: 2399 (51.2\%); 1.0: 1625 (34.7\%) &
0.0: 881 (56.4\%); 1.0: 490 (31.4\%) &
Chi-square &
0.0026 \\
\hline

Alcohol use &
0.0: 2667 (57.0\%); 1.0: 1359 (29.0\%) &
0.0: 897 (57.4\%); 1.0: 472 (30.2\%) &
Chi-square &
0.6495 \\
\hline

Illicit drug use &
0.0: 3645 (77.9\%); 1.0: 378 (8.1\%) &
0.0: 1214 (77.7\%); 1.0: 154 (9.9\%) &
Chi-square &
0.0522 \\
\hline

Occupation &
0.0: 2926 (62.5\%); 1.0: 729 (15.6\%); 2.0: 997 (21.3\%) &
0.0: 963 (61.7\%); 1.0: 261 (16.7\%); 2.0: 318 (20.4\%) &
Chi-square &
0.4640 \\
\hline

First-degree family history of colorectal cancer &
Multiple sparse categories &
Multiple sparse categories &
Chi-square (assumptions violated) &
Not tested \\
\hline

Second-degree family history of colorectal cancer &
0.0: 3973 (84.9\%); 1.0: 42 (0.9\%) &
0.0: 1351 (86.5\%); 1.0: 12 (0.8\%) &
Chi-square &
0.7093 \\
\hline

History of CAD / heart failure &
0.0: 3644 (77.8\%); 1.0: 379 (8.1\%) &
0.0: 1249 (80.0\%); 1.0: 115 (7.4\%) &
Chi-square &
0.2982 \\
\hline

History of diabetes mellitus &
0.0: 2870 (61.3\%); 1.0: 1153 (24.6\%) &
0.0: 999 (64.0\%); 1.0: 365 (23.4\%) &
Chi-square &
0.1890 \\
\hline

History of hypertension &
0.0: 1505 (32.2\%); 1.0: 2521 (53.9\%) &
0.0: 659 (42.2\%); 1.0: 711 (45.5\%) &
Chi-square &
0.0001 \\
\hline

History of inflammatory bowel disease &
0.0: 3974 (84.9\%); 1.0: 54 (1.2\%) &
0.0: 1352 (86.6\%); 1.0: 11 (0.7\%) &
Chi-square &
0.1566 \\
\hline

History of chronic kidney disease &
0.0: 3760 (80.3\%); 1.0: 265 (5.7\%) &
0.0: 1297 (83.0\%); 1.0: 67 (4.3\%) &
Chi-square &
0.0312 \\
\hline

On dialysis &
0.0: 3967 (84.7\%); 1.0: 53 (1.1\%) &
0.0: 1355 (86.7\%); 1.0: 9 (0.6\%) &
Chi-square &
0.0683 \\
\hline

History of cholecystectomy &
0.0: 3904 (83.4\%); 1.0: 78 (1.7\%) &
0.0: 1337 (85.6\%); 1.0: 24 (1.5\%) &
Chi-square &
0.7338 \\
\hline

History of \textit{H. pylori} infection &
0.0: 3929 (83.9\%); 1.0: 110 (2.3\%) &
0.0: 1317 (84.3\%); 1.0: 58 (3.7\%) &
Chi-square &
0.0076 \\
\hline

Aspirin use &
0.0: 3686 (78.7\%); 1.0: 334 (7.1\%) &
0.0: 1245 (79.7\%); 1.0: 120 (7.7\%) &
Chi-square &
0.6183 \\
\hline

Multivitamin use &
0.0: 3901 (83.3\%); 1.0: 118 (2.5\%) &
0.0: 1339 (85.7\%); 1.0: 26 (1.7\%) &
Chi-square &
0.0520 \\
\hline

NSAID use &
0.0: 3744 (80.0\%); 1.0: 271 (5.8\%) &
0.0: 1288 (82.5\%); 1.0: 75 (4.8\%) &
Chi-square &
0.1193 \\
\hline

History of or ongoing chemotherapy/radiation &
0.0: 3922 (83.8\%); 1.0: 96 (2.1\%) &
0.0: 1336 (85.5\%); 1.0: 27 (1.7\%) &
Chi-square &
0.4432 \\
\hline

\end{tabular}

\label{tab:baseline_characteristics}
\end{table}

Categorical variables were compared using chi-square tests when assumptions were met. Missing values were excluded from hypothesis testing but included in descriptive summaries. Statistical testing was not performed for variables with sparse or multi-label categories due to violation of chi-square assumptions.

The mean age was similar across cohorts, although a small but statistically significant difference was observed (internal: 60.16 ± 9.79 years; external: 60.86 ± 10.12 years; p = 0.0018). In the internal cohort, 47.7\% were female and 52.3\% were male, compared with 50.8\% female and 49.2\% male in the external cohort (p = 0.0358).

Most participants were African American, accounting for 84.5\% of the internal cohort and 83.9\% of the external cohort, with smaller proportions of White, Hispanic, Asian, and other racial groups. Absolute differences in racial composition between cohorts were modest.

Body mass index (BMI) distributions were similar between the internal and external cohorts (p = 0.162). Across both cohorts, the largest proportion of patients were overweight (BMI 25-29.9 kg/$\mathrm{m}^2$), and approximately one-third met criteria for obesity (BMI $\ge$ 30 kg/$\mathrm{m}^2$). Smoking prevalence was higher in the internal cohort compared with the external cohort (34.7\% vs. 31.4\%; p = 0.0026), whereas alcohol use did not differ significantly between groups (p = 0.6495).

No statistically significant differences were observed in illicit drug use between the internal and external cohorts (8.1\% vs. 9.9\%, respectively; p = 0.052). Occupation status was similarly distributed across cohorts (p = 0.464). Family history of colorectal cancer, including first- and second-degree relatives, was uncommon in both cohorts; due to sparse and multi-label category distributions, formal statistical testing was not performed for these variables, and comparisons are presented descriptively. 

Regarding comorbid conditions, hypertension was more prevalent in the internal cohort than in the external cohort (53.9\% vs. 45.5\%, p < 0.001), while modest differences were observed for chronic kidney disease (5.7\% vs. 4.3\%, p = 0.031) and Helicobacter pylori infection (2.3\% vs. 3.7\%, p = 0.0076). In contrast, the prevalence of coronary artery disease or heart failure, diabetes mellitus, and inflammatory bowel disease did not differ significantly between cohorts (all $p > 0.05$). Dialysis status and prior cholecystectomy were uncommon and comparable across groups. Similarly, use of aspirin, nonsteroidal anti-inflammatory drugs, multivitamins, and prior or ongoing chemotherapy or radiation therapy showed no statistically significant differences and were of limited clinical magnitude.

Overall, the internal and external cohorts demonstrated broadly similar demographic and clinical profiles, with statistically significant differences largely attributable to the large sample size and small absolute variations. These findings support the appropriateness of using the temporally distinct external cohort for model validation and robustness assessment.

\subsection{Comparison of Machine Learning Models}
We applied multiple machine learning algorithms, including Logistic Regression (LR), Random Forest (RF), Support Vector Machine (SVM), XGBoost, K-Nearest Neighbor (KNN), Neural Network (NN), Decision Trees, and Naïve Bayes (NB). Full performance metrics for internal and external validation, including ROC-AUC, PR-AUC, accuracy, precision, recall, and F1 scores, are presented in Table \ref{tab:model_performance}.

\begin{table}[htbp]
\caption{Performance of machine learning models for high-risk polyp prediction. Internal validation was conducted using the 2015--2022 cohort (n = 4,681), and external validation was conducted using the 2023--2024 cohort (n = 1,562). Performance metrics include ROC-AUC, PR-AUC, accuracy, precision, recall, and F1 score, reported as weighted averages.}
\centering
\small
\renewcommand{\arraystretch}{1.2}
\begin{tabular}{|p{2.55cm}|p{1.5cm}|p{1cm}|p{1cm}|p{1.35cm}|p{1.35cm}|p{0.9cm}|p{0.8cm}|}
\hline
\textbf{Model} &
\textbf{Validation} &
\textbf{ROC-AUC} &
\textbf{PR-AUC} &
\textbf{Accuracy} &
\textbf{Precision (W)} &
\textbf{Recall (W)} &
\textbf{F1 (W)} \\
\hline

\multirow{2}{*}{Neural Network} &
Internal &
0.7764 &
0.7508 &
0.724 &
0.723 &
0.724 &
0.726 \\

 & External &
0.6654 &
0.6600 &
0.656 &
0.656 &
0.656 &
0.656 \\
\hline

\multirow{2}{*}{RF} &
Internal &
0.5898 &
0.5063 &
0.573 &
0.569 &
0.573 &
0.570 \\

 & External &
0.5814 &
0.5655 &
0.569 &
0.570 &
0.569 &
0.565 \\
\hline

\multirow{2}{*}{SVM} &
Internal &
0.5698 &
0.4955 &
0.552 &
0.558 &
0.552 &
0.553 \\

 & External &
0.6317 &
0.6009 &
0.605 &
0.605 &
0.605 &
0.605 \\
\hline

\multirow{2}{*}{KNN} &
Internal &
0.5371 &
0.4542 &
0.529 &
0.537 &
0.529 &
0.531 \\

 & External &
0.5421 &
0.5199 &
0.531 &
0.531 &
0.531 &
0.531 \\
\hline

\multirow{2}{*}{Na\"ive Bayes} &
Internal &
0.5379 &
0.4612 &
0.502 &
0.544 &
0.502 &
0.487 \\

 & External &
0.6152 &
0.6025 &
0.579 &
0.588 &
0.579 &
0.571 \\
\hline

\multirow{2}{*}{XGBoost} &
Internal &
0.5735 &
0.4921 &
0.540 &
0.559 &
0.540 &
0.540 \\

 & External &
0.6012 &
0.5754 &
0.568 &
0.572 &
0.568 &
0.564 \\
\hline

\multirow{2}{*}{Logistic Regression} &
Internal &
0.5698 &
0.4895 &
0.547 &
0.553 &
0.547 &
0.549 \\
 & External &
0.6325 &
0.6035 &
0.597 &
0.598 &
0.597 &
0.597 \\
\hline

\multirow{2}{*}{Decision Tree} &
Internal &
0.5414 &
0.4557 &
0.545 &
0.548 &
0.545 &
0.546 \\

 & External &
0.5248 &
0.5048 &
0.525 &
0.525 &
0.525 &
0.524 \\
\hline

\end{tabular}
\label{tab:model_performance}
\end{table}

\begin{figure}[h]
\centering
\includegraphics[width=1\textwidth]{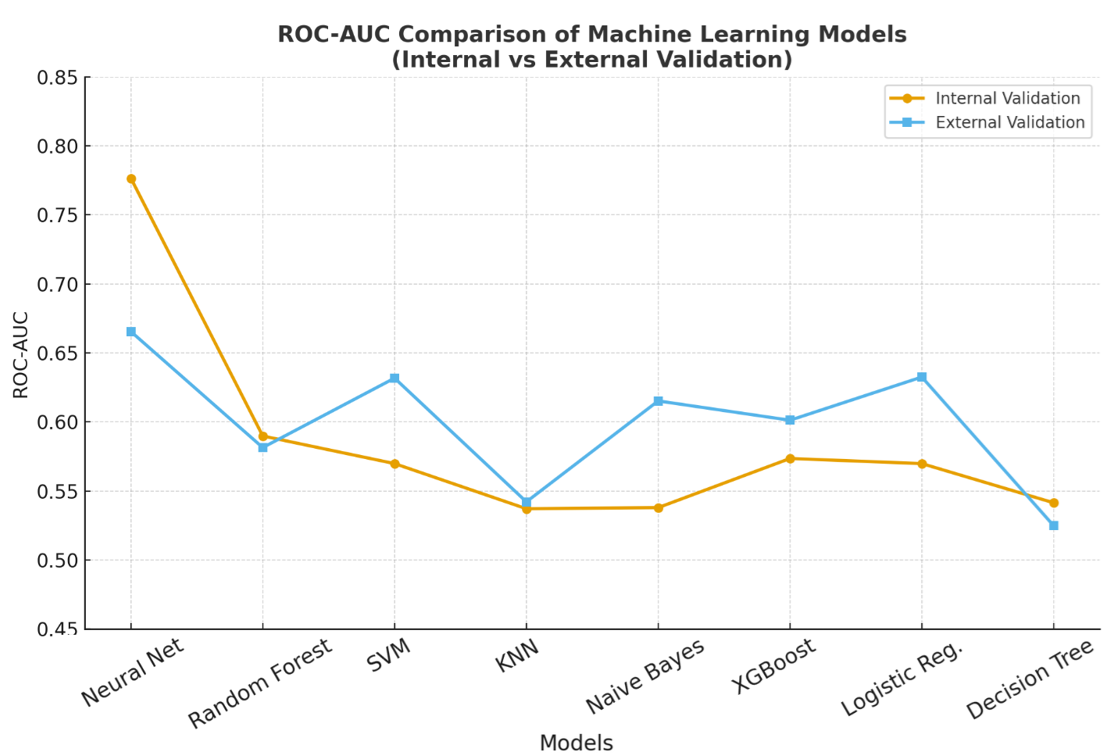}
\caption{Comparison of Model Performance (ROC-AUC) for Internal and External Validation}\label{fig2}
\end{figure}

\begin{figure}[h]
\centering
\includegraphics[width=0.9\textwidth]{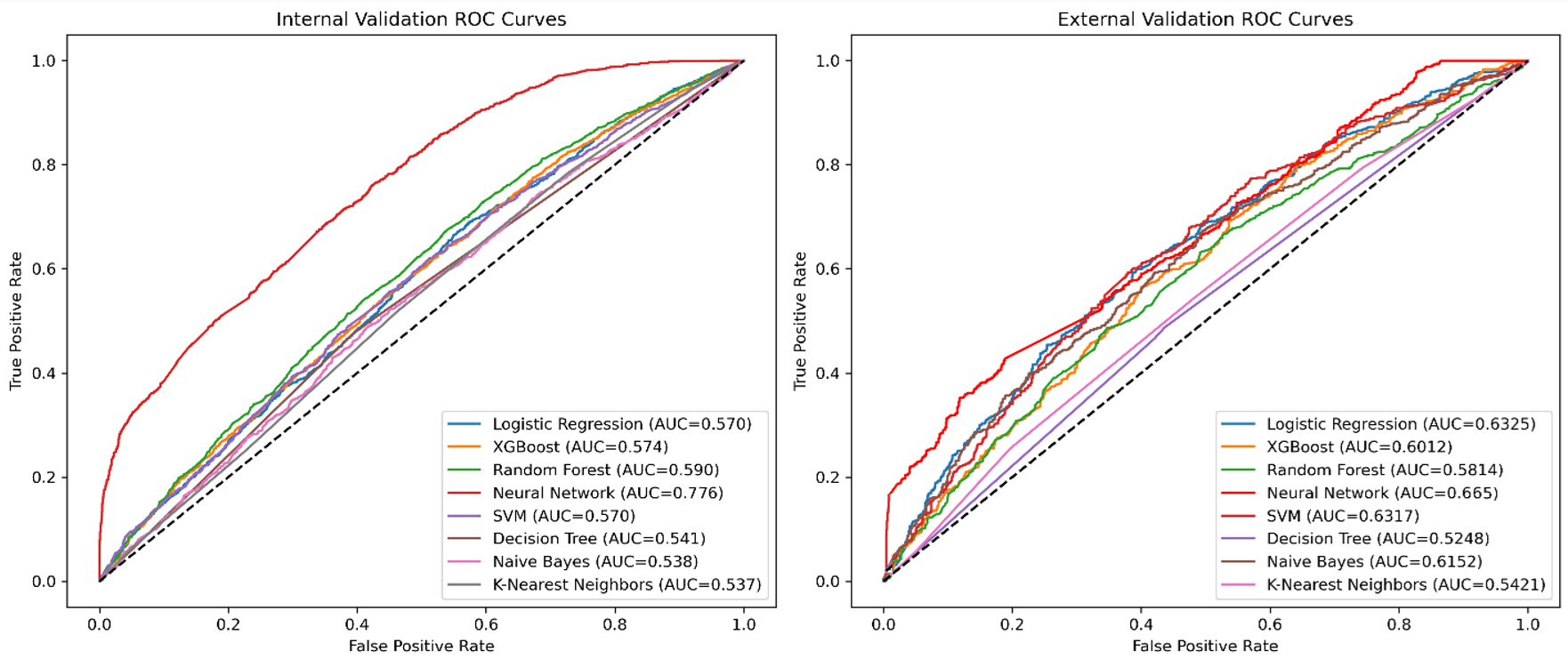}
\caption{ROC Curves for Model Performance in Internal (A) and External (B) Validation Cohorts}\label{fig3}
\end{figure}

Across eight machine learning algorithms, the NN achieved the highest internal validation performance (ROC-AUC 0.776, PR-AUC 0.75,  accuracy  0.72. However, when evaluated on the independent 2023-2024 external cohort, its performance declined (ROC-AUC 0.67, PR-AUC 0.66, accuracy 0.66, F1-score 0.72), suggesting potential overfitting and/or temporal feature drift between cohorts. In contrast, LR, SVM, RF, KNN, Decision Tree, XGBoost, and NB showed lower ROC-AUC values (\textasciitilde{} 0.54-0.59) on the internal cohort but relatively similar or slightly better performance on the external cohort indicating more stable generalization at the cost of internal discrimination. Figure \ref{fig2} summarizes ROC-AUC values for internal and external validation across all models, while Figures \ref{fig3}(A \& B) provides detailed ROC curves for internal and external cohorts, respectively.

\subsection{Model Interpretability}
\begin{figure}[h]
\centering
\includegraphics[width=0.9\textwidth]{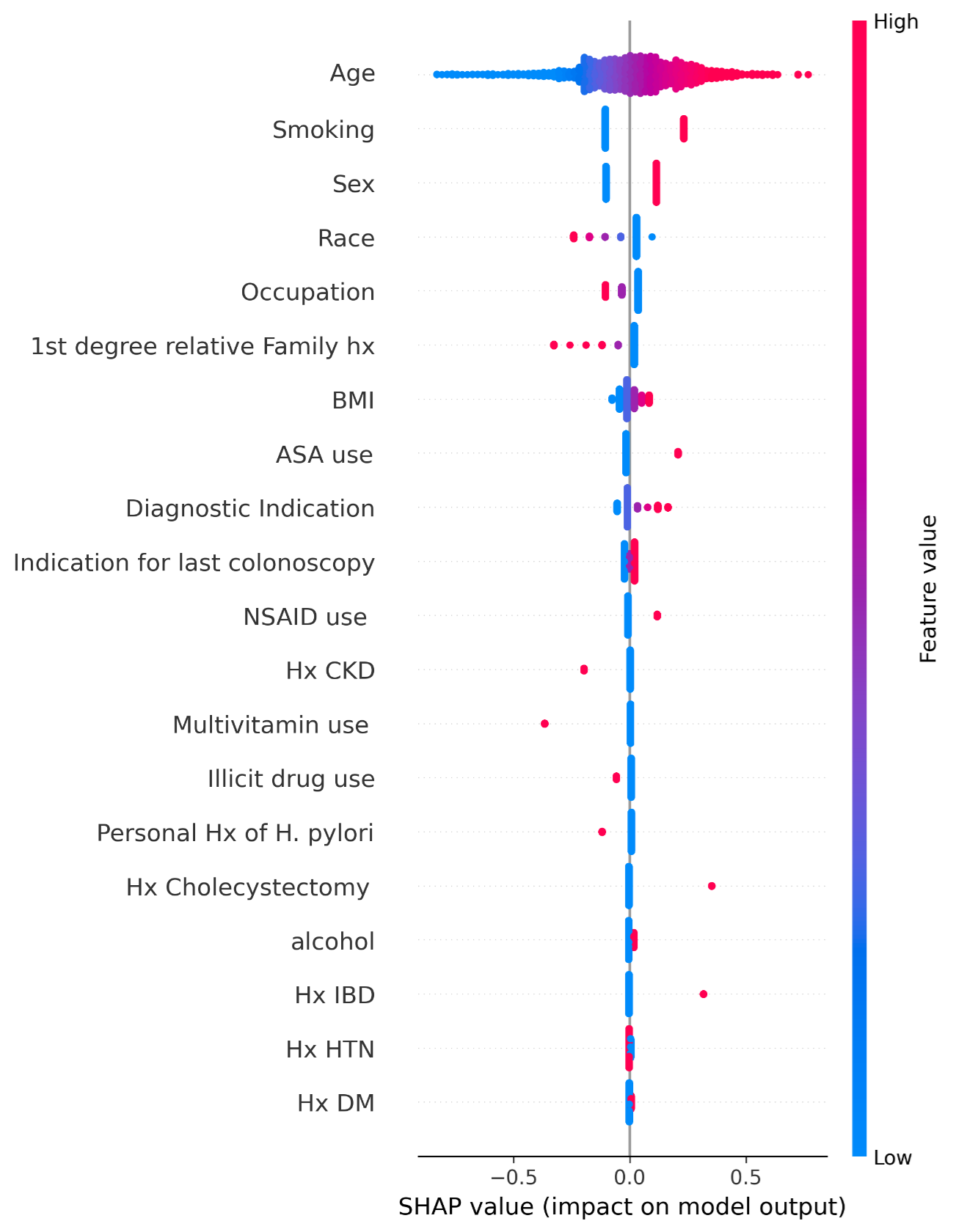}
\caption{SHAP Summary Plot Showing the Most Influential Predictors of High-Risk Polyps}\label{fig4}
\end{figure}

To enhance clinical interpretability, we used SHAP analysis to identify the most influential predictors of high-risk polyps. The top-ranked variables were age, smoking status, sex, occupation, race, and diagnostic indication for colonoscopy (Figure \ref{fig4}). Additional contributors included family history of colorectal cancer in first-degree relatives, BMI, and exposure such as aspirin  (ASA), NSAID use, and alcohol consumption. These findings indicate that traditional clinical risk factors remain dominate drivers of prediction, while sociodemographic variables also contribute meaningfully to model performance.

\subsection{Summary of Findings}
Taken together, these results show that prediction of high-risk polyps using pre-colonoscopy features is feasible. While the neural network model achieved the strongest discrimination in internal validation, its performance decline in the external cohort highlights the risk of overfitting and the influence of temporal or cohort differences. Simpler models demonstrated more robust generalization over time, but at the cost of lower predictive accuracy. The key predictors identified by the models reaffirm the central role of established clinical risk factors while also revealing additional signal from sociodemographic determinants.

\section{Discussion}\label{sec5}
In this study, we developed machine learning models and evaluated their performance using temporal validation on a held-out, later-period cohort to predict high-risk colorectal polyps using only pre-colonoscopy demographic, clinical, and behavioral features.

The findings show that noninvasive, pre-procedural data can provide meaningful discrimination between patients with and without high-risk adeno. Notably, the model performance declined when evaluated on the temporally distinct external cohort (e.g., NN ROC-AUC: 0.78 internal vs. 0.67 external), and most other models performed near chance levels. These findings underscore the limitations of temporal generalizability and highlight the need for future validation in truly independent datasets. Additionally, no decision curve or net benefit analysis was performed, limiting assessment of clinical utility. While modest discrimination may still offer value for triage in resource-limited settings, further work is needed to evaluate practical decision-making thresholds and clinical impact.

The best-performing models achieved ROC-AUC values comparable to or exceeding those reported in prior pre-colonoscopy prediction studies [6-9], which typically relied on limited logistic regression frameworks. The use of modern ML techniques, such as gradient boosting and neural networks, enabled the capture of nonlinear interactions among risk factors such as age, BMI, family history, hypertension, and lifestyle behaviors. Explainability analysis using SHAP provided clinically interpretable insights, confirming that established CRC risk factors remained dominant while revealing new interactions that may warrant further investigation. Importantly, our findings largely confirm previously established colorectal neoplasia risk factors within a predominantly African American urban cohort that has been underrepresented in prior colorectal cancer risk prediction and machine learning studies \cite{jeon_determining_2018, ba_development_2024, sninsky_risk_2022}. Key predictors identified by the models, including age, smoking status, sex, family history of colorectal cancer, and BMI, are consistent with prior risk-stratification literature \cite{yeoh_asia-pacific_2011, jeon_determining_2018, sninsky_risk_2022}. Thus, rather than introducing entirely novel determinants, our study extends existing evidence by demonstrating the feasibility and temporal performance of pre-colonoscopy risk prediction models in a diverse and historically underserved population \cite{ba_development_2024, topol_high-performance_2019}.

Our findings have important clinical implications. A pre-colonoscopy risk prediction model based on routinely available demographic and clinical data could support more personalized screening and surveillance strategies by prioritizing individuals at higher risk for earlier or more frequent colonoscopy, while potentially reducing unnecessary procedures among lower-risk patients. Such an approach aligns with emerging trends in precision screening and health equity, as it avoids reliance on resource-intensive biomarkers or imaging modalities that may not be universally accessible. By leveraging AI-driven analytics applied to diverse clinical datasets, this work provides a foundation for pre-screening tools that may assist gastroenterologists in optimizing colonoscopy utilization, improving early detection of high-risk lesions, and mitigating colorectal cancer disparities in historically underserved populations.

An important consideration is whether machine learning-based risk prediction provides meaningful advantages over conventional questionnaire-based risk scores. Many of the variables identified as important in our models, including age, smoking history, sex, and family history of colorectal cancer, are already included in existing screening questionnaires and clinical risk indices. Therefore, our findings should not be interpreted as replacing traditional approaches, but rather as demonstrating the feasibility of integrating routinely available electronic health record data into automated pre-colonoscopy risk estimation workflows. Compared with static scoring systems, machine learning methods may better accommodate nonlinear interactions, missingness patterns, and higher-dimensional clinical data while reducing manual calculation burden. However, the modest improvement in discrimination observed in this study suggests that the incremental predictive benefit of complex ML approaches over simpler risk scores may currently be limited \cite{yeoh_asia-pacific_2011, jeon_determining_2018, tokutake_artificial_2023}.

\section{Limitations and Future Directions}
Several limitations should be acknowledged. First, this study was conducted using data from a single institution, which may limit generalizability despite the inclusion of an independent temporal external validation set. Second, the study design was retrospective; prspective validation is essential to confirm clinical applicability and assess real-world performance. Third, some variables, such as lifestyle factors and medication use, were self-reported and therefore subject to recall or reporting bias. Additionally, certain variables, such as dietary patterns or medication duration, were unavailable, potentially constraining model performance. Certain variables included in the predictive models, such as illicit drug use, may be considered sensitive and may not be consistently available or uniformly documented across healthcare systems because of privacy, disclosure, or regulatory considerations. Although these variables were obtained from routinely collected clinical records under Institutional Review Board approval and HIPAA-compliant data governance procedures, their incorporation into future clinical prediction tools would require careful consideration of ethical, legal, and implementation issues, including patient privacy, transparency, and potential bias.

Future research should aim to incorporate multi-institutional and multiethnic cohorts, integrate additional data modalities such as genomic, imaging, microbiome, and social determinants for robust ML models. Ultimately, integrating this predictive framework into clinical decision support systems could enable personalized screening strategies and enhance the early detection of colorectal neoplasia.

\section{Clinical Implications}

The proposed predictive framework has the potential to support clinicians in identifying individuals at higher risk of advanced or high-risk polyps before colonoscopy. By leveraging routinely available demographic, behavioral, and clinical features, the model can assist in optimizing colonoscopy scheduling, tailoring surveillance intervals, and prioritizing resource allocation. Integration of such an AI-assisted risk stratification tool into clinical workflows could facilitate earlier detection, improve adherence to screening guidelines, and ultimately reduce the incidence and mortality associated with colorectal cancer. Importantly, the model’s explainability and reliance on pre-colonoscopy data make it practical for real-world adoption in diverse healthcare settings.

\section{Conclusion}

This study demonstrates that machine learning models trained on routinely collected pre-colonoscopy clinical and behavioral data can effectively predict patients at high risk for advanced colorectal polyps. By relying solely on noninvasive features, our approach supports early risk stratification and may improve screening efficiency in diverse health examination populations. The consistent performance across internal and external datasets underscores the robustness and potential clinical utility of the proposed framework. Future work will focus on prospective validation, integration with electronic health records, and model adaptation for multi-institutional use to further enhance precision colorectal cancer prevention and early detection strategies.

\backmatter

\begin{appendices}

\section{TRIPOD-AI Checklist}\label{secA1}

\renewcommand{\arraystretch}{1.2}
\begin{longtable}{|p{3.16cm}|p{2.6cm}|p{8cm}|}
\caption{Supplementary Table S1. TRIPOD-AI Checklist}
\label{tab:tripod_ai} \\
\hline
\textbf{TRIPOD-AI Item} & \textbf{Section / Page} & \textbf{Response} \\
\hline
\endfirsthead

\hline
\textbf{TRIPOD-AI Item} & \textbf{Section / Page} & \textbf{Response} \\
\hline
\endhead

\hline
\endfoot

\hline
\endlastfoot

\multicolumn{3}{|l|}{\textbf{Title and Abstract}} \\
\hline
1. Title & Title page & The title clearly identifies the study as development and temporal external validation of machine learning models for predicting high-risk colorectal polyps using routinely available pre-colonoscopy clinical data. \\
\hline
2. Abstract & Abstract & The abstract summarizes objectives, study design, data source, cohort sizes, predictors, modeling approaches, internal and temporal external validation, performance metrics, and key conclusions in accordance with TRIPOD-AI guidance. \\
\hline

\multicolumn{3}{|l|}{\textbf{Introduction}} \\
\hline
3a. Background and rationale & Introduction & The clinical rationale for developing pre-colonoscopy risk prediction models is described, including limitations of current screening strategies and the potential role of machine learning using non-invasive EHR data. \\
\hline
3b. Objectives & Introduction & The study objective is to develop and temporally externally validate machine learning models to predict high-risk colorectal polyps using pre-procedural clinical features. \\
\hline

\multicolumn{3}{|l|}{\textbf{Methods}} \\
\hline
4. Study design & Methods & A retrospective observational cohort study was conducted using electronic health record data from Howard University Hospital. \\
\hline
5a. Data source & Methods & Data were obtained from institutional electronic health records, including demographic, clinical, behavioral, medication, and family history variables collected prior to colonoscopy. \\
\hline
5b. Study dates & Methods & Colonoscopies performed between January 2015 and December 2024 were included. Internal (2015--2022) and temporal external (2023--2024) cohorts were defined by colonoscopy date. \\
\hline
6a. Eligibility criteria & Methods & All adult patients undergoing colonoscopy during the study period were eligible. No exclusions were based on outcome status. \\
\hline
6b. Outcome definition & Methods & The outcome was presence of high-risk colorectal polyps as determined by post-colonoscopy pathology, defined according to established clinical criteria. \\
\hline
7a. Predictors & Methods & Candidate predictors included demographics, comorbidities, lifestyle behaviors, medication use, family history, and colonoscopy indication, all available prior to the procedure. \\
\hline
7b. Predictor assessment & Methods & All predictors were extracted from structured EHR fields and assessed prior to outcome determination. \\
\hline
8. Sample size & Methods & No formal sample size calculation was performed. All eligible patients meeting inclusion criteria during the study period were included to maximize statistical power and model stability. \\
\hline
9. Missing data & Methods & Missing values were excluded from statistical testing for baseline comparisons. Missingness was reported descriptively but was not included as a category in hypothesis testing. No imputation was performed for baseline descriptive analyses. \\
\hline
10a. Statistical analysis methods & Methods & Baseline characteristics were compared between cohorts using Mann--Whitney U tests for continuous variables and chi-square or Fisher's exact tests for categorical variables when assumptions were met. Variables with sparse or multi-label categories were summarized descriptively without formal hypothesis testing. \\
\hline
10b. Model development & Methods & Multiple machine learning models were trained using the internal cohort, with standardized preprocessing and cross-validation. \\
\hline
10c. Model performance & Methods & Model discrimination was assessed using ROC-AUC, with calibration and additional metrics reported as appropriate. \\
\hline
10d. Validation & Methods & Temporal external validation was performed using an independent cohort defined by later calendar years. \\
\hline
11. Risk groups & Methods & No predefined risk thresholds were used; predicted probabilities were analyzed continuously. \\
\hline

\multicolumn{3}{|l|}{\textbf{Results}} \\
\hline
13a. Participant flow & Results & A total of 6,243 patients were included, with 4,681 in the internal cohort and 1,562 in the temporal external cohort. \\
\hline
13b. Participant characteristics & Results & Baseline demographic, clinical, and behavioral characteristics are reported in Table~1, with cautious interpretation of statistical comparisons. \\
\hline
14a. Model development results & Results & Model performance during internal validation is reported for all candidate algorithms. \\
\hline
14b. Model specification & Results & Final model architectures, feature sets, and training procedures are described. \\
\hline
15a. Model performance & Results & Discrimination performance is reported for internal and temporal external validation cohorts. \\
\hline
15b. Model comparison & Results & Comparative performance across machine learning models is presented. \\
\hline

\multicolumn{3}{|l|}{\textbf{Discussion}} \\
\hline
18. Limitations & Discussion & Limitations include retrospective design, single-center data, modest external discrimination, and potential residual confounding. \\
\hline
19a. Interpretation & Discussion & Results are interpreted in the context of existing literature, emphasizing generalizability and clinical relevance rather than statistical significance of baseline differences. \\
\hline
19b. Clinical use & Discussion & The potential role of pre-colonoscopy risk stratification to support personalized screening strategies is discussed, with acknowledgment of the need for further validation. \\
\hline

\multicolumn{3}{|l|}{\textbf{Other Information}} \\
\hline
20. Ethics & Methods & Institutional Review Board approval was obtained, and informed consent was waived due to the retrospective nature of the study. \\
\hline

\end{longtable}




\end{appendices}


\bibliography{sn-bibliography}

\end{document}